\documentclass[10pt,twocolumn,letterpaper]{article}

\usepackage[pagenumbers]{cvpr} 

\definecolor{cvprblue}{rgb}{0.21,0.49,0.74}
\usepackage[pagebackref,breaklinks,colorlinks,allcolors=cvprblue]{hyperref}
\usepackage{subcaption}
\usepackage{bm}
\usepackage{multirow}
\usepackage{adjustbox}
\usepackage{xspace}
\usepackage{pifont}
\usepackage{wrapfig}
\newcommand{\cmark}{\ding{51}}%
\newcommand{\xmark}{\ding{55}}%
\newcommand{\ieno}{\textit{i}.\textit{e}.}
\newcommand{\egno}{\textit{e}.\textit{g}.} 
\newcommand{\dataname}{\emph{MotionBank}\xspace}

\def\paperID{*****} 
\def\confName{CVPR}
\def\confYear{2025}

\title{MotionBank: A Large-scale Video Motion Benchmark with Disentangled Rule-based Annotations}

\newcommand*{\affmark}[1][*]{\textsuperscript{#1}}

\author{%
Liang Xu\affmark[1,2] \quad
Shaoyang Hua\affmark[1,3] \quad
Zili Lin\affmark[1,2] \quad
Yifan Liu\affmark[2] \quad
Feipeng Ma\affmark[3] \quad
Yichao Yan\affmark[2] \\
Xin Jin\affmark[1]\thanks{Corresponding author} \quad
Xiaokang Yang\affmark[2] \quad
Wenjun Zeng\affmark[1] \quad
\vspace{0.7em} \\
\affmark[1]{Eastern Institute of Technology, Ningbo} \quad
\affmark[2]{Shanghai Jiao Tong University} \\
\affmark[3]{University of Science and Technology of China} \\
\vspace{-0.5em}
}

\begin{document}
\maketitle

\begin{abstract}

In this paper, we tackle the problem of how to build and benchmark a large motion model (LMM). 
The ultimate goal of LMM is to serve as a foundation model for versatile motion-related tasks, \egno, human motion generation, with interpretability and generalizability. 
Though advanced, recent LMM-related works are still limited by small-scale motion data and costly text descriptions.
Besides, previous motion benchmarks primarily focus on pure body movements, neglecting the ubiquitous motions in context, \ieno, humans interacting with humans, objects, and scenes.
To address these limitations, we consolidate large-scale video action datasets as knowledge banks to build MotionBank, which comprises \textbf{13} video action datasets, \textbf{1.24M} motion sequences, and \textbf{132.9M} frames of natural and diverse human motions.
Different from laboratory-captured motions, in-the-wild human-centric videos contain abundant motions in context.
To facilitate better motion text alignment, we also meticulously devise a motion caption generation algorithm to \textit{automatically} produce rule-based, unbiased, and disentangled text descriptions via the kinematic characteristics for each motion.
Extensive experiments show that our MotionBank is beneficial for general motion-related tasks of human motion generation, motion in-context generation, and motion understanding.
Video motions together with the rule-based text annotations could serve as an efficient alternative for larger LMMs.
Our dataset, codes, and benchmark will be publicly available at \href{https://github.com/liangxuy/MotionBank}{https://github.com/liangxuy/MotionBank}.

\end{abstract}

\section{Introduction}
Understanding human activities has been fundamental and widely studied for decades~\cite{carreira2017quo,zhu2020comprehensive}. Compared to human-centric images or videos, motions show the advantages of the efficient and essential human activity representation. Besides plenty of works on image/video-based human pose estimation, human motion generation~\cite{humanml3d,mdm} has also attracted much attention with numerous real-world applications for digital humans, AR/VR, gaming, and human-robot interactions.

Building a practical human motion generation model with interpretability and generalizability is imperative. However, modeling human motions is challenging for that the underlying human activities are intricate, diverse, long-tailed, with various interactions.
Previous attempts for generalizable human motion generation can be summed up in three perspectives:
1) \textbf{Motion-based solutions} tend to enrich the motion data with in-the-wild static poses~\cite{make_an_animation}, synthesized motion~\cite{cai2021playing,xu2023actformer,black2023bedlam} or integrating multiple motion datasets~\cite{motion-x,LMM2024}.
2) \textbf{Text-based solutions} typically achieve better text motion alignment with richer and fine-grained descriptions~\cite{action-gpt,shi2023generating,he2023semanticboost,motion-x} such as directions, orientations, hand statuses and body parts augmented by LLMs.
3) \textbf{Pre-trained models based solutions}~\cite{motiongpt,zhang2024motiongpt,zhou2023avatargpt,wu2024motionllm,chen2024motionllm} leverage the foundation large language models~\cite{radford2018improving,radford2019language,brown2020language,ouyang2022training,touvron2023llama,liu2023llava}, and successfully build a unified framework for motion generation, resulting in large motion models (LMMs).

\begin{figure*}[t]
  \centering
  \includegraphics[width=1.0\linewidth]{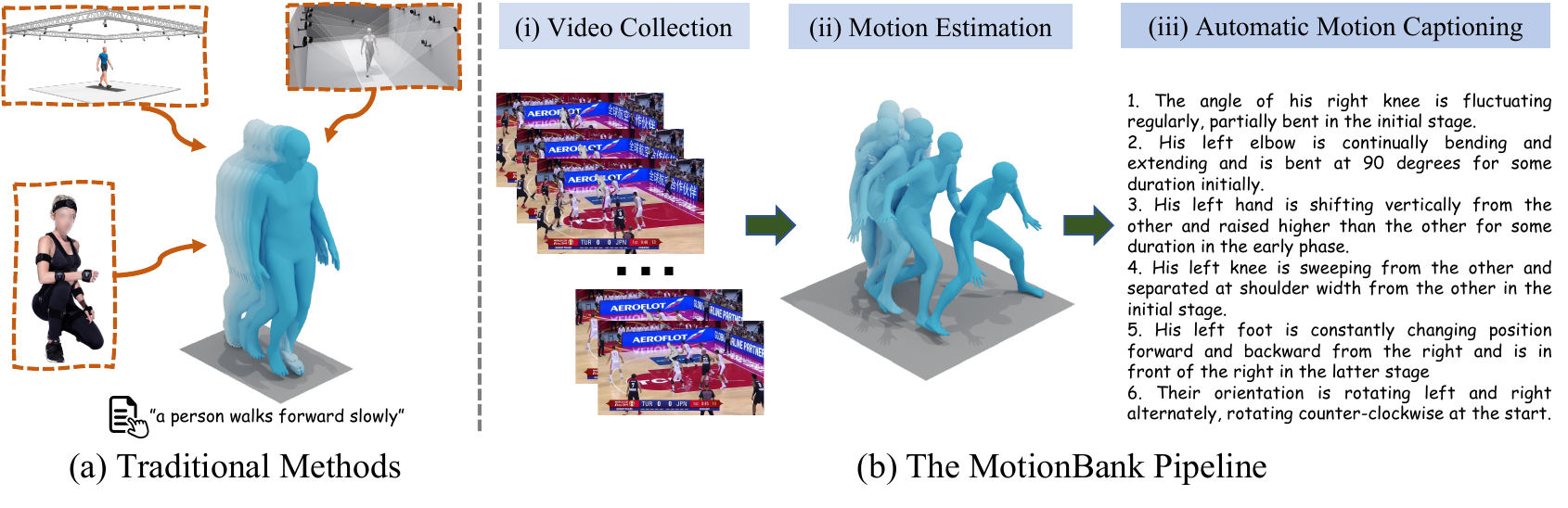}
  \caption{\textbf{Difference illustration of our proposed MotionBank.} (a) Previous datasets are collected from optical/inertial motion capture (MoCap) systems, multi-view cameras, and manual text annotations. (b) For \dataname, we collect vast in-the-wild human-centric videos from the public and extract human motion from them. We also devise an algorithm to {automatically} generate the rule-based, fine-grained, and disentangled motion captions as the corresponding text annotations.}
  \label{fig:teaser}
\end{figure*}

Despite advances of LMMs from the perspectives of motion data, text annotation, and pre-trained models, we find that previous works are still unsatisfactory for practical applications compared to the large models based on languages~\cite{gpt3,gpt4}, images~\cite{radford2021learning,li2022blip} and videos. 
For the motion data, previous LMMs primarily rely on motion capture data in laboratory scenes with limited diversity and negligible in-context motions, \ieno, lacking human interaction with humans, objects, and scenes. 
For the text annotation, the manual labeling of detailed text descriptions is also tedious and expensive, and the LLM-produced texts also might be unreliable for intricate motion patterns.
Therefore, here comes the question, \textit{``How to comprehensively and efficiently benchmark a large motion model?''}. Considering that the ultimate goal of LMM is to serve as a foundation model for modeling versatile human activities, we answer this question from three aspects: 1) a large-scale motion dataset sampling from the realistic distribution of human daily activities is desired; 2) the corresponding text descriptions should be fine-grained, while the annotation cost is also tolerant; 3) the advanced ability of LMMs to empower various downstream motion tasks in context is also required.

In this paper, we turn to large-scale human-centric video action datasets~\cite{liu2009recognizing,niebles2010modeling,kliper2011action,kuehne2011hmdb,soomro2012ucf101,zhang2013actemes,charades,kinetics400,gu2018ava,nturgbd120,chung2021haa500,li2021uav,li2021multisports} as motion knowledge banks for a new benchmark as in~\cref{fig:teaser}, which has at least the following advantages: 1) \textbf{Larger Quantity.} In the pyramid of data acquisition, 4D motions are much more scarce than videos, images, and languages. With the rapid development of video pose estimation techniques~\cite{cao2017realtime,fang2022alphapose,jin2020whole,xu2022zoomnas,sun2023trace,ye2023decoupling,shin2023wham}, especially for robustness on in-the-wild videos and capturing global trajectories, extracting millions of motion data from videos is feasible. 2) \textbf{More Natural and Diverse.} In-the-wild human-centric videos are mainly collected from our daily lives, sufficient amount of them will have the potential to get close to the real distribution of human activities. Extreme motions such as ``diving'', and ``triple jump'' are also covered. 3) \textbf{In-context Motion Involved.} Human interactions with the surrounding environments in the video action datasets are ubiquitous and beneficial to generalizing LMMs.
We use 13 widely-adopted video action datasets and extract the SMPL~\cite{smpl} parameters to form \dataname, which comprises \textbf{1.24M} sequences, \textbf{1225.7} hours and \textbf{132.9M} frames of motion data.
Each motion sequence contains a rule-based text, which results in \textbf{1.24M} texts.
To the best of our knowledge, \dataname is currently the largest text-motion dataset in ~\cref{tab:dataset_comp}.

With the processed motion data, the corresponding text annotations are also indispensable building blocks for better motion text mapping. However, it is impractical to label each motion caption manually. Inspired by previous works on automatic pose/motion description synthesis~\cite{bourdev2009poselets,pons2014posebits,kim2021fixmypose,delmas2022posescript,fieraru2021aifit,yazdian2023motionscript,he2023semanticboost,liu2023bridging}, we manage to devise an automatic motion caption generation algorithm based on the concept of posecodes~\cite{delmas2022posescript}. Posecodes are low-level, generic rules based on the kinematic characteristics of static poses.
Different from images or videos represented by \texttt{incomprehensible} pixel values, motions are represented as \texttt{understandable} low-level coordinates for machines. For example, posecodes can be defined as ``\textit{the elbow is slightly/partially/completely bent}'' for different angle intervals by calculating the angle composed of the coordinates of <wrist, elbow, shoulder>. We extend the descriptions of static posecodes to temporal maintenances/changes of posecodes with timing and duration depictions. Since the process is completely automatic and based on joint-level pose states, we can obtain rule-based, unbiased, and disentangled text annotations efficiently.

\begin{table*}[t]
\centering
\resizebox{1\linewidth}{!} {%
    \begin{tabular}{lcccccc|ccc}
    \toprule
    Dataset & Year & Motions & Frames & Hours & Texts & Source & HHI & HOI & HSI \\ \midrule
    KIT-ML~\cite{plappert2016kit} & 2016 & 3.91K & 245K & 11.2  & 6.28K & MoCap & \xmark & \xmark & \xmark \\ 
    AMASS~\cite{mahmood2019amass} & 2019 & 11.27K & 20M & 40.0 & 0 & MoCap & \xmark & \xmark & \xmark \\ 
    BABEL~\cite{punnakkal2021babel} & 2021 & 13.22K & 7M & 43.5 & 91.41K & MoCap & \xmark & \xmark & \xmark \\
    HumanML3D~\cite{humanml3d} & 2022 & 14.62K & 2M & 28.6 & 44.97K & MoCap & \xmark & \xmark & \xmark \\
    Motion-X~\cite{motion-x} & 2023 & 81.08K & 15.6M & 144.2 & 81.08K & MoCap\&Videos & \xmark & \xmark & \xmark \\ \midrule
    \textbf{MotionBank} & 2024 & \textbf{1.24M} & \textbf{132.9M} & \textbf{1225.7} & \textbf{1.24M} & Videos & \cmark & \cmark & \cmark\\
    \bottomrule
    \end{tabular}}
\caption{\textbf{Datasets Comparison.} We compare \dataname with previous text-motion datasets. Our crowdsourcing dataset presents currently the largest motion knowledge base, together with a new level of diversity for human-human (HHI), human-object (HOI), and human-scene (HSI) interactions.}
\label{tab:dataset_comp}
\end{table*}

With the paired motions and rule-based text annotations, we first conduct extensive comparisons to validate the advantages of \dataname. 
Then, we follow previous attempts to propose a uniform motion-language framework based on large language models~\cite{touvron2023llama}. Specifically, we first quantize the motion data into discrete indices of a learned codebook with a vector quantized variational autoencoder (VQ-VAE~\cite{vq-vae}) model, and then evaluate the single-person human motion generation model on the widely adopted HumanML3D~\cite{humanml3d} dataset. To further benchmark the adaptation ability for in-context motion generation, we also evaluate the model on the human motion generation for human-human~\cite{xu2023inter}, human-object~\cite{behave} and human-scene~\cite{wang2022humanise} interactions.
Experimental results show that our \dataname is beneficial for general motion-related tasks of human motion generation and motion in-context generation. The contributions of this paper can be summarized as follows:
\begin{itemize}
    \item We propose a new large-scale video motion dataset \dataname with \textbf{1.24M} motion sequences and \textbf{132.9M} frames of natural and diverse human motions.
    \item We explore an efficient alternative benchmark towards large motion models with video motions and rule-based disentangled text annotations.
    \item Extensive experiments show the benefits of \dataname for versatile motion-related tasks.
\end{itemize}

\section{Related Work}

\textbf{Video Action Recognition.} 
Human action recognition is a significant and long-standing research problem with numerous efforts on building large-scale datasets, including in-the-wild videos~\cite{schuldt2004recognizing,liu2009recognizing,kuehne2011hmdb,kliper2011action,soomro2012ucf101,zhang2013actemes,caba2015activitynet,kinetics400,gu2018ava,zhao2019hacs,mit,diba2020large,monfort2021multi,li2021uav}, sports videos~\cite{karpathy2014large,ibrahim2016hierarchical,shao2020finegym}, indoor videos~\cite{charades,nturgbd120,das2019toyota,goyal2017something,damen2018scaling}. In this paper, we take large-scale video action datasets as abundant motion knowledge banks. We believe that extracting knowledge from videos is promising for 3D models, even 4D motions.


\textbf{Human Motion Generation.}
Human motions can be viewed as an efficient and compact human activity representation. Many datasets~\cite{nturgbd120,humanact12,uestc,punnakkal2021babel,plappert2016kit,humanml3d,motion-x,li2021learn,Valle-Perez2021Transflower,li2023finedance} and generative methods~\cite{csgn,actor,petrovich22temos,action2motion,cervantes2022implicit,mdm,mofusion,motiondiffuse,commdm,motiongpt,zhang2023remodiffuse,zhang2024motiongpt,LMM2024,chen2024motionllm} are proposed in recent years.
Different from \textit{Motion-X}~\cite{motion-x}, our \dataname obtains all the motion data from videos and our obtained captions are unbiased, disentangled, and fine-grained. From the benchmark perspective, our focus lies in diverse motion-related tasks, ranging from single-person human motion generation to motion in-context generations. Zhang~\etal~\cite{LMM2024} also consolidates 16 motion datasets for LMMs, yet it mainly focuses on indoor MoCap motion data and a unified human motion generation model with various input signals. Recent works~\cite{motiongpt,zhang2024motiongpt,zhou2023avatargpt,wu2024motionllm} achieve great improvements by injecting motion into large language models~\cite{radford2018improving,radford2019language,brown2020language,ouyang2022training,touvron2023llama,liu2023llava}. However, all these works are trained on limited data and neglect the in-context motion generation abilities.

\textbf{Motion in Context.} In-context human motion is defined as humans interacting with the surrounding humans~\cite{van2011umpm,sbu_kinect,ng2020you2me,nturgbd120,chi3d,expi,hi4d,interhuman,xu2023inter}, objects~\cite{GRAB:2020,behave,huang2022intercap,Jiang_2023_ICCV_chairs,fan2023arctic,li2023objectmotionguided} and scenes~\cite{cao2020long,savva2016pigraphs,hassan2019resolving,hassan2021stochastic,wang2022humanise,huang2022capturing,guzov23ireplica,araujo2023circle}.
The acquisition of large-scale 4D human in-context data is quite expensive, which hinders the development of these domains.
Xu~\etal~\cite{xu2024interdreamer} \textit{firstly} tackle the problem of zero-shot human object interaction by decoupling the interaction semantics and dynamics, where the dynamics (\ieno, the pure body movements) can be generated by off-the-shelf human motion generation model and the object trajectories can be synthesized based on the human motion and physical constraints. The foundation model trained on abundant in-context motions in our \dataname also reveals the effectiveness of HOI decomposition in improving the overall generation quality.

\textbf{Automatic Motion Captioning.}
The low-level kinematic characteristics of human motion are extensively analyzed and leveraged for automatic text annotations~\cite{bourdev2009poselets,pons2014posebits,kim2021fixmypose,fieraru2021aifit,delmas2022posescript,yazdian2023motionscript,delmas2023posefix,liu2023bridging}. Pons-Moll~\etal~\cite{pons2014posebits} propose \textit{posebits} as boolean geometric relationships between body parts. Delmas~\etal~\cite{delmas2022posescript} manage to generate still pose descriptions based on \textit{posecodes}. Yazdian~\etal~\cite{yazdian2023motionscript} generate natural language descriptions for 3D motions. Uniquely, to ensure motion pattern diversity and consistency, we focus on \textit{``posecodes''} changes or stationary \textit{``posecodes''} for long times. Descriptions of starting time and duration are also incorporated for clear illustration.

\textbf{Multimodal Large Language Models.}
Recent multimodal large language models have achieved promising performance in various domains, including image understanding~\cite{liu2023llava,chen2023shikra,wang2023visionllm,zhang2023llavar,li2023monkey,zhu2023minigpt,zhang2023internlmXcomposer}, image  generation~\cite{pan2024kosmosg,jin2023unified, ge2023planting_seed}, and video understanding~\cite{lin2023videollava,zhang2023videollama,maaz2023videochatgpt,song2023moviechat,chen2023videollm}.
Further research advances LLMs from 2D to 3D-related tasks. 3D-LLM~\cite{hong20233dllm} collects 1M 3D-language data and train the LLM with a 3D feature extractor for diverse 3D-related tasks.
MotionGPT~\cite{zhang2024motiongpt} proposes to fine-tune the LLM as a general-purpose motion generator. 
Jiang~\etal~\cite{motiongpt} employs discrete vector quantization to transfer human motion into motion tokens and performs language modeling on both motion and text.
However, these works on motions are limited by the insufficient amount of data, which makes them cannot fully unleash the LLM potential. To address this issue, we propose a large-scale motion-text pairs dataset, \dataname, which comprises 1.24M motion sequences, with the corresponding fine-grained text annotations.

\begin{table}[t]
\centering
\resizebox{\linewidth}{!}{
  \makeatletter\def\@captype{table}\makeatother
  \begin{tabular}{lccccc}
      \toprule
          Dataset & Clips & Source & Classes & Motions & Frames \\ \midrule
          YouTube Action~\cite{liu2009recognizing} & 1,600 & Wild & 11 & 1,684 & 132,781 \\ 
          Olympic Sports~\cite{niebles2010modeling} & 783 & Sports & 16 & 3,215 & 342,591 \\ 
          ASLAN~\cite{kliper2011action} & 3,697 & Wild & 432 & 5,834 & 507,133 \\ 
          HMDB51~\cite{kuehne2011hmdb} & 6,766 & Wild & 51 & 9,521 & 703,954 \\ 
          UCF101~\cite{soomro2012ucf101} & 13,320 & Wild & 101 & 23,734 & 2,170,230 \\ 
          Penn Action~\cite{zhang2013actemes} & 2,326 & Wild & 15 & 3,245 & 212,676 \\ 
          Charades~\cite{charades} & 9,848 & Indoor & 157 & 26,636 & 7,296,693 \\ 
          Kinetics-400~\cite{kinetics400} & 240,436 & Wild & 400 & 848,347 & 85,632,244 \\ 
          AVA~\cite{gu2018ava} & 430 & Wild & 80 & 96,890 & 15,336,180 \\ 
          NTU RGB+D 120~\cite{nturgbd120} & 114,480 & Indoor & 120 & 46,436 & 3,037,510 \\ 
          HAA500~\cite{chung2021haa500} & 10,000 & Wild & 500 & 10,548	& 652,178 \\ 
          UAV-Human~\cite{li2021uav} & 22,476 & Wild & 155 & 40,334 & 6,355,724 \\ 
          MultiSports~\cite{li2021multisports} & 3,200 & Sports & 4 & 119,381 & 10,557,767 \\ \midrule
          MotionBank & 429,362 & - & 1739 & 1,235,805 & 132,937,661 \\ \bottomrule
      \end{tabular}
  }
  \captionof{table}{\small \textbf{Dataset Statistics.} We collect 13 video action datasets with extracted motion data to form \dataname.}
  \label{tab:all_datasets}
\end{table}

\begin{figure}[t]
  \centering
  \includegraphics[width=6.2cm]{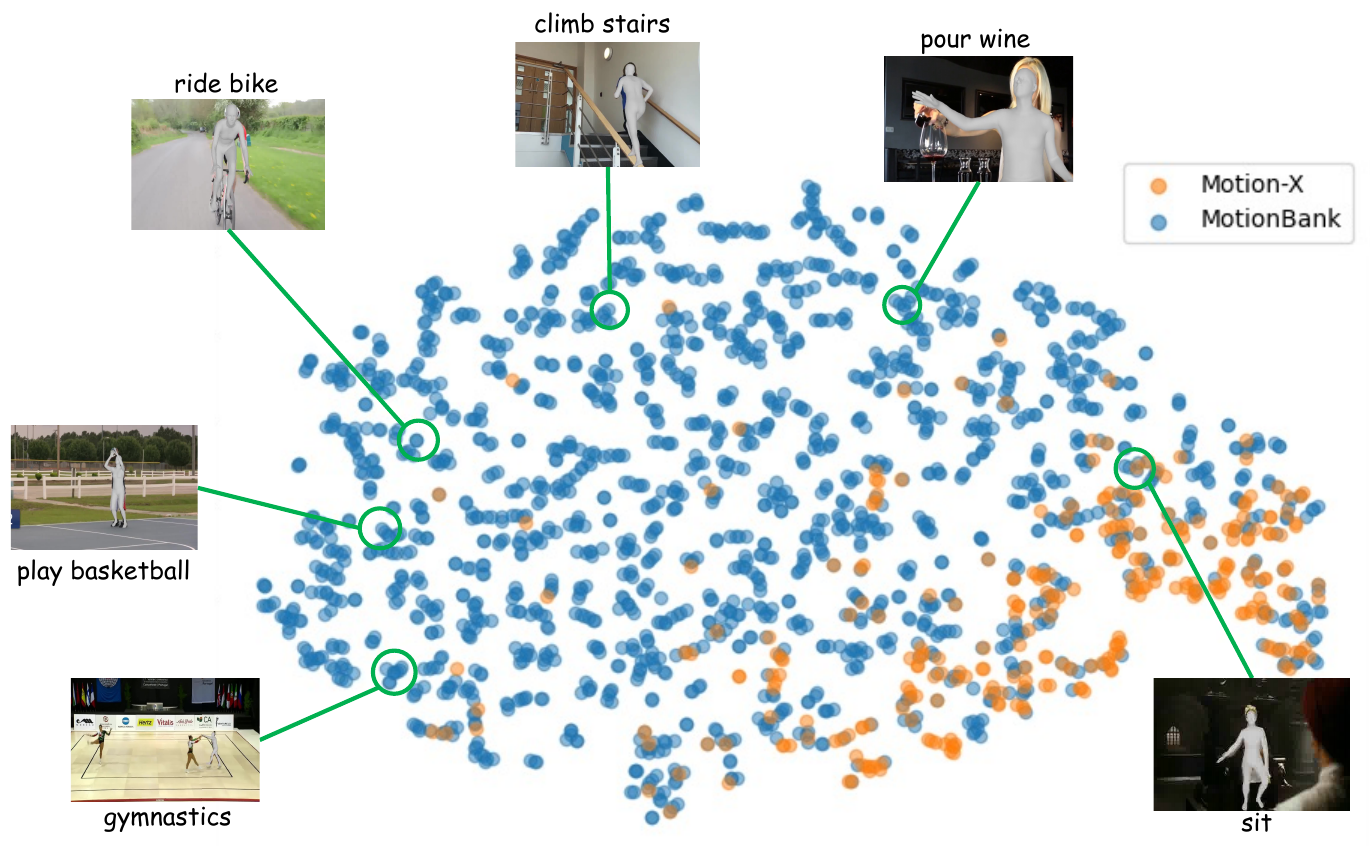}
  \caption{\small Visualization of semantic space distributions between \textit{Motion-X} and \dataname.}
  \label{fig:cluster} 
\end{figure}

\section{The Proposed MotionBank}

\subsection{The Basic Construction of MotionBank}
As aforementioned, we turn to human-centric video action datasets as motion knowledge banks for the large quantity of natural, diverse, and in-context motions.
We collect 13 video action datasets with different kinds of in-the-wild, indoor, and sports datasets as depicted in~\cref{tab:all_datasets}, resulting in \textbf{429.3K} video clips.~\cref{fig:cluster} also show that \dataname contains richer motion semantics.
We remove the duplicate action classes in~\cref{tab:all_datasets}.
Most of the adopted datasets are collected based on in-the-wild videos~\cite{liu2009recognizing,kuehne2011hmdb,soomro2012ucf101,zhang2013actemes,kinetics400,chung2021haa500,li2021uav}, like K400~\cite{kinetics400}, which is a large-scale dataset with 400 human action classes on the YouTube website. AVA~\cite{gu2018ava} is built on 430 different movie scenes with natural daily humans interacting with the surrounding environments. Specifically, we also gather some sports videos~\cite{niebles2010modeling,li2021multisports} for the richness of the dataset on extreme motions, human-human interactions, and group motions. Charades~\cite{charades} and NTU RGB+D 120~\cite{nturgbd120} are two video action datasets with abundant daily indoor activities and human-human interactions~\cite{nturgbd120}.

\begin{figure*}[t]
  \centering
  \includegraphics[width=1.0\linewidth]{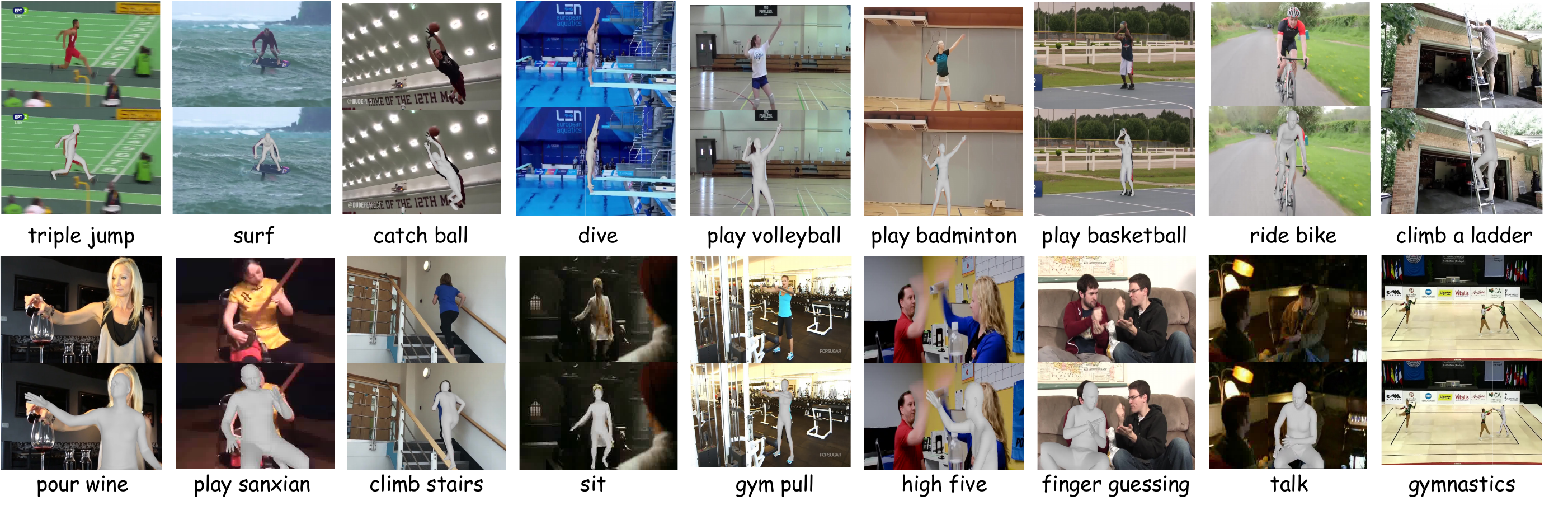}
  \caption{\textbf{Visualization of the video motion data of MotionBank.} The crowdsourcing 13 video action datasets contain abundant 1) In-the-wild daily activities, 2) Sports; 3) Natural and diverse human in context motions, \ieno, human-human, object, scene interactions.}
  \label{fig:motion_vis}
\end{figure*}

After comprehensive comparisons, we adopt the WHAM~\cite{shin2023wham} to extract the SMPL parameters of the collected videos. Empirically, WHAM generalizes well to unseen videos and the regressed 3D human motion is in the world rather than the camera coordinate system, which is indispensable for modeling the root translations of the movements. We provide some visualization results of the video scenes and the extracted SMPL parameters in~\cref{fig:motion_vis}. Note that, if a scene contains multiple people, we keep all the motion tracklets of different people. For example, given \textit{``one person pushes the other, and the other falls back''}, we keep both the motion of \textit{``pushing''} and \textit{``falling back''} as two separate motions to enrich our \dataname with ample in-content motion patterns.

\subsection{Automatic Motion Captioning}
\begin{center}
    ``\emph{What You See Is What You Get.}''
\end{center}

Different from images or videos that are stored by \texttt{incomprehensible} pixel values, the motion can be represented as \texttt{understandable} low-level coordinates, \ieno, \texttt{getting the text descriptions upon seeing the motion}. Pons-Moll~\etal~\cite{pons2014posebits} advocated the concept of \textit{posebits} as boolean geometric relationships between body parts, such as \textit{``Is right hand above the hips''} according to the relative position between the right hand and hips, and \textit{``Is right knee bent''} according to the angle of the knee. Delmas~\etal~\cite{delmas2022posescript} extended and proposed the low-level pose information as \textit{posecodes} to describe the relation between a specific set of joints, such as the \textit{``angle posecodes''} to describe how a body part \textit{``bents''} at a given joint. The angle is discretized into several intervals with each corresponding to a text category as \textit{``slightly/partially/$\dots$/completely bent''}. 
Details of the rules are presented in the supplementary materials. Next, we first introduce the concepts of \textit{posecodes}~\cite{delmas2022posescript} for single poses and then elaborate the process of aggregating them into motion descriptions.
\textit{Posecodes} are introduced as the kinematic relation between a specific set of joints with the following five categories:

\textbf{Angle.} To depict the degree of flexion/extension of a specific body joint, such as elbows and legs.

\textbf{Distance.} To measure the $\mathit{L}$2-distance between two body parts, \egno, two hands.

\textbf{Relative Position.} To describe how two keypoints are positioned for a given axis. For the \textit{x}-axis from body's right to left horizontally, the possible categories could be \{``at the right of'', ``at the left of'', ``x-ignored''\}, where ``ignored'' will be ineligible for final descriptions.

\begin{figure}[t]
  \centering
  \includegraphics[width=1.0\linewidth]{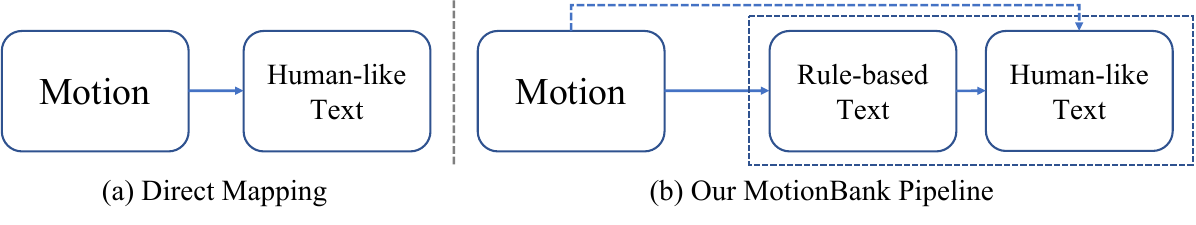}
  \caption{\textbf{The pipeline of text motion alignment.} (a) Previous methods adopt the direct mapping between motions and human-like texts. (b) Our \dataname take the rule-based texts as a bridge to narrow the gap between motions and human-like texts via fine-tuning.}
  \label{fig:pipeline}
\end{figure}

\textbf{Pitch \& Roll.} To evaluate the verticality or horizontality of a bone defined by two keypoints, such as the elbow and wrist defining the forearm.

\textbf{Ground-Contact.} To indicate whether the contact occurs between a keypoint and the ground.

Similarly, to describe the statuses of the holistic orientation and translations of the motion sequence, we define the two extra \textit{posecodes} categories as:

\textbf{Orientation.} To specify the body root orientation along three axes relative to the first frame. For example, the orientation around the \textit{x}-axis could be \{``lie backward'', ``lean backward'', ``x-ignored'', ``lean forward'', ``lie forward''\} for different rotation angles.

\textbf{Translation.} To characterize the locations of the body along three axes relative to the first frame. For example, the translation along the \textit{x}-axis could be \{``move left'', ``x-ignored'', ``move right''\}.

\begin{figure}[t]
  \centering
  \includegraphics[width=0.50\textwidth]{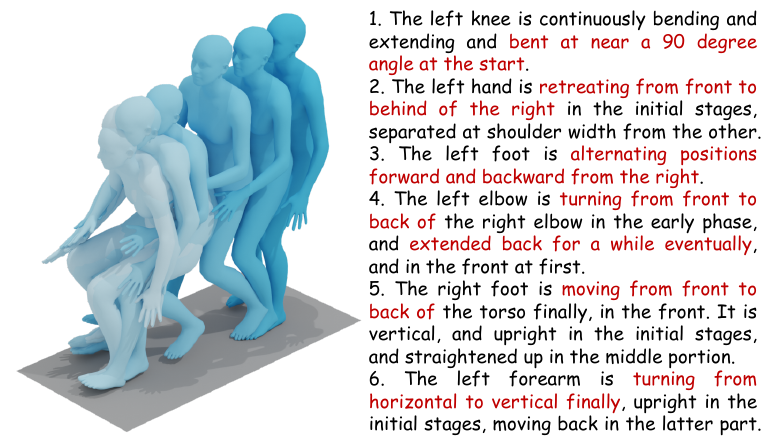}
  \caption{\small \textbf{Examples of the generated motion captions.} In this example from Multisports~\cite{li2021multisports}, the man steps back from a half-squat pose. Our generated results can correctly capture the dynamics and semantics.}
  \label{fig:motionscript_res}
\end{figure}

With the above definitions, we illustrate the process of extending to automatic motion captioning. After comprehensive explorations, we devise a simple yet effective framework based on the \textit{posecodes} of all the frames.
We define the temporal tracklet for each \textit{posecode} as \textit{motioncode} as in~\cite{yazdian2023motionscript}.
The details can be found in the supplementary material.
Our insight is that (a) one \textit{motioncode} is eligible \textbf{if and only if} one of its \textit{posecode} is eligible.
(b) We only focus on significant \textbf{changes} in \textit{posecode} statuses and \textbf{long-duration} stationary statuses.
For better temporal awareness of the described motions, we incorporate the temporal attributes of start timing, such as \{``initially'', ``in the middle'', ``ultimately''\} and durations, such as \{``for a short time'', ``for a while'', ``for a long time'', ``for the whole period''\} for each \textit{motioncode} into the posecodes. 
For each \textit{motioncode}, diverse expression methods are designed to introduce variability into the descriptions. 
We also add some random skip strategies to avoid redundancy. 
As illustrated in~\cref{fig:pipeline}, we believe that the rule-based texts can serve as intermediate cornerstones to narrow the gap between motion and human-like texts.
More caption results can be found in~\cref{fig:motionscript_res} and supplementary materials.


All in all, compared with human-like texts, our rule-based texts have several advantages: 1) \textbf{Fine-grained}, with human part level descriptions; 2) \textbf{Authentic}, unbiased without human-induced errors; 3) \textbf{Cost friendly}, without human labor and training models.

\section{Large Motion Model}

The model architecture is inspired by MotionGPT~\cite{motiongpt,zhang2024motiongpt} to propose a unified motion-language framework based on large language models~\cite{radford2018improving,radford2019language,brown2020language,ouyang2022training,touvron2023llama} by motion quantization (~\cref{sec:motion_vq}), motion-language pretraining (~\cref{sec:motion_text_pret}) and adaptive tuning for downstream tasks (~\cref{sec:downstream_tuning}).

\subsection{Motion Quantization}
\label{sec:motion_vq}
Given a motion $X=\{{x}^i\}_{i=1}^{M}$ with $M$ frames, we train a motion tokenizer $\mathcal{V}$ based on VQ-VAE~\cite{vq-vae,guo2022tm2t,t2m-gpt} with a motion encoder $\mathcal{E}$ and a motion decoder $\mathcal{D}$. 
The encoder $\mathcal{E}$ first extracts the latent feature $\hat{z}^{1:L}=\mathcal{E}(x^{1:M})$ with 1D convolutions along the temporal dimension, where $L=M/l$ and $l$ is the temporal downsampling rate. $M$ is the length of the whole sequence and $L$ is the length of the latent feature after the encoder. Then, we quantize the latent features $\hat{z}^{1:L}$ into a codebook $C=\{{c}^k\}_{k=1}^{K} \in \mathbb{R}^{K\times d}$, where $K$ is the size of the codebook and $d$ is the dimension for each codebook entry. The quantization process is to find the index of the most similar element of the codebook $C$ as:
\begin{equation}
    z_i:={\arg \min }_{c_k \in C}\left\|\hat{z_i}-c_k\right\|_2.
\end{equation}
After quantization, the decoder $\mathcal{D}$ projects the learned discrete representation $z^{1:L}=\{{z}^i\}_{i=1}^{L}$ back to the original motion data as $\hat{x}^{1:M}$.
We follow~\cite{guo2022tm2t,t2m-gpt,motiongpt} to train the motion tokenizer with three loss functions as $ \mathcal{L}_\mathcal{V} = \mathcal{L}_{r} + \mathcal{L}_{e} + \mathcal{L}_{c}\label{eq:loss:vq}$, where $\mathcal{L}_{r}$ is the reconstruction loss, $\mathcal{L}_{e}$ is the embedding loss and $\mathcal{L}_{c}$ is the commitment loss. We follow~\cite{motiongpt} to utilize $L$1 smooth loss and velocity regularization for $\mathcal{L}_r$. During training, exponential moving average (EMA) and codebook reset~\cite{razavi2019generating} are also adopted for better codebook utilization. More details of the losses and training are provided in the supplementary.

\subsection{Motion-Language Pre-training}
\label{sec:motion_text_pret}
With the trained motion tokenizer, we convert the continuous human motions into discrete motion tokens as $x^{1:M} \mapsto z^{1:L}$, so that we can take motions as languages and apply the vocabulary embedding in language models~\cite{kudo2018sentencepiece,raffel2023exploring,ouyang2022training} to the motion tokens. Following~\cite{motiongpt}, we learn the motion and text jointly by combining the text vocabulary and the motion vocabulary. With the unified text-motion vocabulary, we can take the motion data as special texts. We learn the mapping between the language and motions in an auto-regressive manner. During the inference time, the discrete motion tokens are projected back to the motion data with decoder $\mathcal{D}$ of the tokenizer.

\subsection{Adaptive Tuning}
\label{sec:downstream_tuning}
Our large motion model trained on large-scale video motion dataset can serve as a foundation model for versatile human-related works. For in-context motion generation such as human-object interaction, InterDreamer~\cite{xu2024interdreamer} decompose the problem into pure body motion generation and then integrating with plausible object manipulation results. Thus, our model can boost the performance for the first stage of HOI generation.
With newcome paired text-motion data for downstream tasks, we can directly finetune the parameters of the trained models. For the new motion data, it share the previous trained motion tokenizer to obtain the discrete codes. The alignment between the human-like text descriptions and our rule-based texts as shown in~\cref{fig:pipeline} can also be achieved in this stage.

\begin{figure*}[t]
  \centering
  \includegraphics[width=1.0\linewidth]{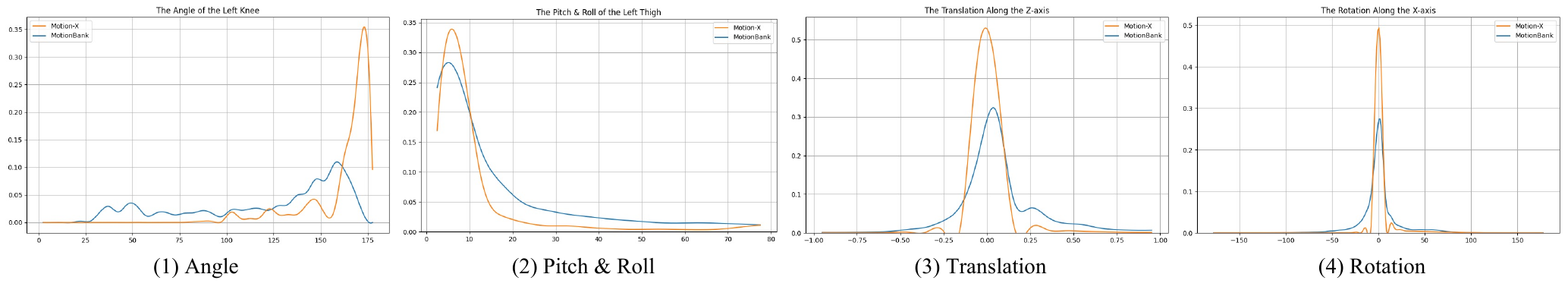}
  \caption{\textbf{Motion space distribution visualizations between Motion-X and MotionBank}, which show that our \dataname has wider and more balanced motion distributions, thus covering more diverse motion patterns. We randomly select 10,000 samples of each dataset for visualization.}
  \label{fig:posecodes}
\end{figure*}

\begin{table*}[t]
\centering
\resizebox{\linewidth}{!}{
\begin{tabular}{lccccccc}
\toprule
\multirow{2}{*}{Methods}&\multicolumn{3}{c}{R Precision$\uparrow$}&\multicolumn{1}{c}{\multirow{2}{*}{FID$\downarrow$}}&\multirow{2}{*}{MMDist$\downarrow$}&\multirow{2}{*}{Diversity$\rightarrow$}&\multirow{2}{*}{MModality$\uparrow$}\\\cmidrule(lr){2-4}
&\multicolumn{1}{c}{Top-1}&\multicolumn{1}{c}{Top-2}&\multicolumn{1}{c}{Top-3}&\multicolumn{1}{c}{}&&&\\\midrule
Real & $0.511^{\pm.003}$ & $0.703^{\pm.003}$ & $0.797^{\pm.002}$ & $0.002^{\pm.000}$ & $2.974^{\pm.008}$ & $9.503^{\pm.065}$ & \multicolumn{1}{c}{-} \\ \midrule
T2M~\cite{humanml3d}& $0.457^{\pm.002}$ & $0.639^{\pm.003}$ & $0.740^{\pm.003}$ & $1.067^{\pm.002}$ & $3.340^{\pm.008}$ & $9.188^{\pm.002}$ & $2.090^{\pm.083}$ \\ \midrule  
T2M~\cite{humanml3d}+Ours & $\mathbf{0.483^{\pm.003}}$ & $\mathbf{0.677^{\pm.003}}$ & $\mathbf{0.776^{\pm.002}}$ & $\mathbf{0.451^{\pm.003}}$ & $\mathbf{3.126^{\pm.011}}$ & $\mathbf{9.452^{\pm.082}}$ & $\mathbf{2.235^{\pm.068}}$\\ \bottomrule
\end{tabular}}
\caption{\textbf{Single-person human motion generation.} Comparison of text-to-motion generation on HumanML3D dataset. $\pm$ indicates 95\% confidence interval and $\rightarrow$ means the closer the better.}
\label{tab:humanml3d}
\end{table*}

\section{Experiment}

Firstly, we measure the diversity of \dataname by visualizing the distrubutions of different categories of \textit{posecodes} in~\cref{sec:exp_motion_diversity}. Then, we design experiments to show the benefits brought by our \dataname for downstream tasks in~\cref{sec:exp_downstream}. Finally, we provide an insightful experiment to show that our proposed rule-based texts allow for plausible human motion generation in~\cref{sec:exp_disentangle}. Due to space limitation, the experiments on the large motion models, extra in-context motion generation, motion understanding, and more implementation details are presented in the supplementary materials.

\subsection{Evaluation of the Motion Diversity}
\label{sec:exp_motion_diversity}
The visualization of the semantic space distributions between \textit{Motion-X} and \dataname is presented in~\cref{fig:cluster}. Here, we provide more visualization results on the comparisons of the motion space distributions. We randomly sample 10,000 motions from the two datasets and visualize several typical \textit{posecodes} in~\cref{fig:posecodes}.
Generally, the \textit{Motion-X} dataset are more long-tailed and biased towards regular motion statuses, such as ``standing'' (from (1) and (2)), ``staying still without translations'' (from (3)) and less extreme rotations (from (4)).
Take the sub-graph of ``angle of the left knee'' as an example, our \dataname contains more bent statuses of the knee.
In conclusion, our \dataname contains broader semantic spaces together with more diverse motion spaces than previous. Please refer to more visualizations in supplementary.

\subsection{Evaluation of Benefits for Downstream Tasks}
\label{sec:exp_downstream}
In this section, we evaluate the generalization ability of the proposed \dataname on the downstream motion-related generative tasks, \ieno, single human motion generation and human-object interaction. For a fair comparison, we train the original T2M~\cite{humanml3d} model with same parameters on our proposed \dataname dataset and then finetune the downstream tasks based on the trained model.

\noindent{\textbf{Single Human Motion Generation.}} We finetune our trained model on the HumanML3D~\cite{humanml3d} dataset, with 14,616 motion sequences and 44,970 text descriptions.
Similar to previous works, we employ the R Precision to quantify the accuracy of top-1, top-2 and top-3 retrieval from 31 randomly mismatched descriptions against the ground-truth description, the Frechet Inception Distance (FID)~\cite{fid} to gauge the distribution distance between real and generated samples, the diversity as a metric for assessing latent variance. Besides, multimodality (MModality) and MultiModal distance (MMDist) are harnessed to appraise the diversity of motions generated from the same text, and to compute the latent discrepancy between generated motions and texts, respectively. Detailed evaluation metric descriptions are provided in supplementary materials for space limitation.
The comparison results presented in~\cref{tab:humanml3d} yield improvements for all metrics and show that our \dataname is beneficial for downstream single human motion generation task.

\noindent{\textbf{Human-Scene Interaction.}} We adopt the widely used BEHAVE~\cite{behave} dataset for validation. Similarly to InterDreamer~\cite{xu2024interdreamer}, we only generate the body movement part of the HOI sequence. After finetuning, significant improvements are achieved in~\cref{tab:behave} to reveal that our \dataname can boost the downstream in-context motion generation to a great extent.

\begin{table*}[t]
\centering
\resizebox{\linewidth}{!}{
\begin{tabular}{lccccccc}
\toprule
\multirow{2}{*}{Methods}&\multicolumn{3}{c}{R Precision$\uparrow$}&\multicolumn{1}{c}{\multirow{2}{*}{FID$\downarrow$}}&\multirow{2}{*}{MMDist$\downarrow$}&\multirow{2}{*}{Diversity$\rightarrow$}&\multirow{2}{*}{MModality$\uparrow$}\\\cmidrule(lr){2-4}
&\multicolumn{1}{c}{Top-1}&\multicolumn{1}{c}{Top-2}&\multicolumn{1}{c}{Top-3}&\multicolumn{1}{c}{}&&&\\\midrule
Real & $0.219^{\pm.008}$ & $0.376^{\pm.010}$ & $0.498^{\pm.009}$ & $0.102^{\pm.005}$ & $3.709^{\pm.021}$ & $6.132^{\pm.063}$ & \multicolumn{1}{c}{-} \\ \midrule
T2M~\cite{humanml3d} & $0.132^{\pm.015}$ & $0.249^{\pm.033}$ & $0.353^{\pm.014}$ & $16.078^{\pm.708}$ & $5.728^{\pm.026}$ & $5.892^{\pm.092}$ & $1.794^{\pm.051}$\\ \midrule
T2M~\cite{humanml3d}+Ours & $\mathbf{0.182^{\pm.008}}$ & $\mathbf{0.316^{\pm.007}}$ & $\mathbf{0.432^{\pm.009}}$ & $\mathbf{1.517^{\pm.063}}$ & $\mathbf{4.692^{\pm.019}}$ & $\mathbf{5.912^{\pm.073}}$ & $\mathbf{2.137^{\pm.084}}$\\ \bottomrule
\end{tabular}}
\caption{\textbf{In context motion generation for human-object interactions.} Comparison of text-to-motion generation on the body motion generation of BEHAVE~\cite{behave} dataset.}
\label{tab:behave}
\end{table*}

\noindent \textbf{Implementation Details.} 
The dimension of the pose vector of \dataname is 263 as in~\cite{humanml3d}. The word features are 300-d embeddings obtained via GloVe~\cite{pennington2014glove} and the maximum text length is set to 300.
We firstly train the T2M~\cite{humanml3d} model on our \dataname and then, we initialize the motion encoder and decoder with the learned parameters and re-train the models for HumanML3D~\cite{humanml3d} and BEHAVE~\cite{behave}. 
All experiments are conducted on one Nvidia A100 GPU with 80 GB memory.

\begin{figure*}[t]
  \centering
  \includegraphics[width=1.0\linewidth]{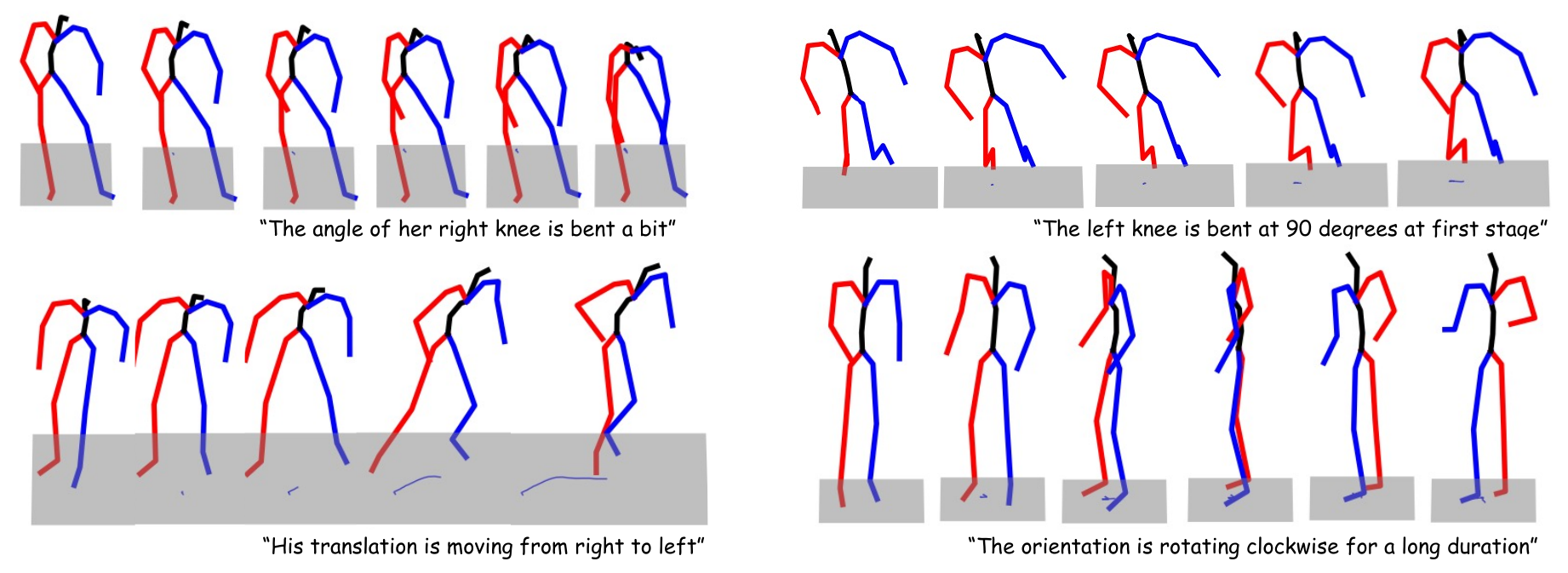}
  \caption{\textbf{Controllable Rule-Text Based Motion Generation.} The generated motion sequences and part of the corresponding rule-based text inputs achieve good alignments.}
  \label{fig:gen_vis}
\end{figure*}

\subsection{Controllable Rule-Text Based Motion Generation}
\label{sec:exp_disentangle}
With the trained T2M~\cite{humanml3d} model on our \dataname, we generate the motions given the prompts of our rule-based descriptions.
The visualization results depicted in~\cref{fig:gen_vis} show that with our rule-based texts, the generated motion sequences can also align well to the text prompts.
This is insightful and promising towards disentangled and controllable human motion synthesis.
More visualization of the generated results and video results will be provided in the supplementary materials.

\section{Conclusion}

In this paper, we explore to extract the abundant motion knowledge from human-centric video action datasets with a motion captioning algorithm to automatically generate rule-based, unbiased and disentangled text descriptions based on the kinematic features for each motion. 
Following this pipeline, we produce \dataname, the currently largest text motion data with \textbf{1.24M} samples and \textbf{132.9M} frames from 13 video action datasets. Statistics results and visualizations show that our \dataname could complement for existing MoCap data that is primarily collected from indoor scenes.
We believe that our pipeline is potential as an efficient alternative towards larger motion models for more open-vocabulary and practical settings.

\noindent{\textbf{Limitation and Future Work.}}
Our setting has some limitations for the following perspectives: 
1) Our trained large motion model only serves as a pre-trained model encoding abundant motion knowledge and the alignment with rule-based texts, thus small-scale motion datasets with manual descriptions are still needed for fine-tuning to bridge the gap between rule-based texts and human-like descriptions.
2) We adopt the SMPL parameters without dexterous hand movements and facial expressions, limited by the capabilities of pose estimation algorithms.
In our future works, we would like to consolidate more human-centric datasets to enhance \dataname and apply LLMs to re-write and optimize our automatically generated texts.
We are also interested in the better mapping between the rule-based and human-like motion descriptions and rule-based motion editing. 
In this work, we only keep the human motion yet neglect the contexts. We would like to directly learning the human context interactions from videos in the future.

\noindent{\textbf{Broader Impact.}}
Generalizable human motion generation is less explored yet significant for AR/VR, gaming and video generation, where LMMs are the most promising pathway. We believe that our proposed \dataname can serve as an effective and efficient alternative to foster future researches on large motion models.
Some potential negative societal impacts may include that our model can be used to synthesize realistic human motion or human interacting with the environments, which may lead to misinformation generation or be abused on inappropriate occasions.

\noindent{\textbf{Acknowledgement.}} 
This work was supported in part by NSFC 62302246 and ZJNSFC under Grant LQ23F010008, and supported by High Performance Computing Center at Eastern Institute of Technology, Ningbo, and Ningbo Institute of Digital Twin.

\clearpage
\appendix

\twocolumn[
\begin{center}
\Large{\bf MotionBank: A Large-scale Video Motion Benchmark with Disentangled Rule-based Annotations \\ **Appendix**}
\end{center}
]

\renewcommand{\thefigure}{\Alph{figure}}
\renewcommand{\thetable}{\Alph{table}}

\section{Details of Large Motion Models}

\subsection{Training Losses of VQ-VAE}

We follow~\cite{guo2022tm2t,t2m-gpt,motiongpt} to train the motion tokenizer with three loss functions as:
\begin{equation}
    \mathcal{L}_\mathcal{V} = \mathcal{L}_{r} + \mathcal{L}_{e} + \mathcal{L}_{c} 
    = \mathcal{L}_{r} + \underbrace{||\mathit{sg}[Z] - \hat{Z}||_2}_{\mathcal{L}_{e}} + \beta \underbrace{||Z - \mathit{sg}[\hat{Z}]||_2}_{\mathcal{L}_{c}}
\end{equation}
where $\beta$ is a hyper-parameter, $sg$ indicates the stop-gradient operator. The reconstruction loss is defined as the $L$1 smooth loss and an additional regularization term on the velocity.

\subsection{Evaluation Metrics}

We first train a motion feature extractor together with a text feature extractor in a contrastive paradigm to align the features of texts and motions as in~\cite{humanml3d}.
All the evaluations are run for 20 times (except MModality for 5 times) and we report the averaged results with the confidence interval at 95\%. 

\begin{itemize}
    \item \textbf{R Precision}: which measures the accuracy of retrieving the ground-truth description from 32 randomly mismatched descriptions, with 1 ground-truth and 31 randomly selected negative descriptions. We report the Top-1, Top-2 and Top-3 accuracy of the retrieval results;
    \item \textbf{FID}~\cite{fid}: The features are extracted from the generated motions and the real motions. Then the FID is calculated between the feature distribution of the generated motions and the distribution of the real motions;
    \item \textbf{Diversity}: which measures the variance of the generated motions across all action categories. Given the motion feature vectors of generated motions and real motions as \{$v_1, \cdots, v_{S_d}$\} and \{$v_1^\prime, \cdots, v_{S_d}^\prime$\}, the diversity is defined as $Diversity=\frac{1}{S_d}\sum_{i=1}^{S_d}||v_i-v_i^\prime||_2$. $S_d=200$ in our experiments;
    \item \textbf{MMDist}: which calculates the latent distance between generated motions and texts as the average Euclidean distances between each text feature and the generated motion feature from the given text;
    \item \textbf{Multi-modality}: which measures how much the generated motions diversify with each action type. Given a set of motions with $C$ action types, for $c$-th action, we randomly sample two subsets with size $S_l$, and then extract the feature vectors as \{$v_{c,1}, \cdots, v_{c, S_l}$\} and \{$v_{c,1}^\prime, \cdots, v_{c,S_l}^\prime$\}, the multimodality is defined as $Multimod.=\frac{1}{C\times S_l}\sum_{c=1}^C\sum_{i=1}^{S_l}||v_{c,i}-v_{c,i}^\prime||_2$. $S_l=20$ in our experiments.
\end{itemize}

\section{Details of Automatic Motion Captioning}

\subsection{Posecode Definitions}
We provide the details of the \textit{posecode} definitions in~\cref{tab:code_cat}, which is borrowed from the original paper of~\cite{delmas2022posescript}.
For the \textit{posecodes} of ``Orientation'' and ``Translation'', we consider the global orientation of SMPL~\cite{smpl} and the root translations, respectively.

The conditions of the \textit{posecode} categorizations are presented in~\cref{tab:elementary_posecodes_thresholds}. Based on the original definitions from~\cite{delmas2022posescript}, we replenish the ``orientation'' and ``translation'' conditions. The ``orientation'' is measured in degrees along different axes, and ``translation'' is measured in meters along different axes.

\begin{table*}[t]
    \centering
    \resizebox{\textwidth}{!}{%
    \begin{tabular}{ccccc}
    \toprule
    \textbf{Angle} & \textbf{Distance} & \textbf{Relative Position} & \textbf{Relative Orientation} & \textbf{Ground-contact} \\ \hline
    L-knee & L-elbow vs R-elbow & L-shoulder vs R-shoulder (YZ) & L-hip vs L-knee & L-knee \\
    R-knee & L-hand vs R-hand & L-elbow vs R-elbow (YZ) & R-hip vs R-knee & R-knee \\
    L-elbow & L-knee vs R-knee & L-hand vs R-hand (XYZ) & L-knee vs L-ankle & L-foot \\
    R-elbow & L-foot vs R-foot & neck vs pelvis (XZ) & R-knee vs R-ankle & R-foot \\
     & L-hand vs L-shoulder & L-ankle vs neck (Y) & L-shoulder vs L-elbow & L-hand\\
     & L-hand vs R-shoulder & R-ankle vs neck (Y) & R-shoulder vs R-elbow & R-hand\\
     & R-hand vs L-shoulder & L-hip vs L-knee (Y) & L-elbow vs L-wrist & \\
     & R-hand vs R-shoulder & R-hip vs R-knee (Y) & R-elbow vs R-wrist & \\
     & L-hand vs R-elbow & L-hand vs L-shoulder (XY) & pelvis vs L-shoulder & \\
     & R-hand vs L-elbow & R-hand vs R-shoulder (XY) & pelvis vs R-shoulder & \\
     & L-hand vs L-knee & L-foot vs L-hip (XY) & pelvis vs neck & \\
     & L-hand vs R-knee & R-foot vs R-hip (XY) & L-hand vs R-hand & \\
     & R-hand vs L-knee & L-wrist vs neck (Y) & L-foot vs R-foot & \\
     & R-hand vs R-knee & R-wrist vs neck (Y) &  & \\
     & L-hand vs L-foot & L-hand vs L-hip (Y) &  & \\
     & L-hand vs R-foot & R-hand vs R-hip (Y) &  & \\
     & R-hand vs L-foot & L-hand vs torso (Z) &  & \\
     & R-hand vs R-foot & R-hand vs torso (Z) &  & \\
     & L-hand vs L-ankle&L-foot vs torso (Z) & &\\
     & L-hand vs R-ankle&R-foot vs torso (Z) & &\\
     & R-hand vs L-ankle&L-knee vs R-knee (YZ) & &\\
     & R-hand vs R-ankle&L-foot vs R-foot (XYZ)\\
    \bottomrule 
    \end{tabular}
    }
    \vspace{0.1cm}
    \caption{\textbf{Pose Code Categories.} We provide the keypoints involved in the five types of the \textit{posecodes}. We denote Left and Right as `L' and `R' respectively. The axes considered for relative positions are noted in parentheses for ``Relative Position''.}
    \vspace{-0.3cm}
    \label{tab:code_cat}
\end{table*}

\begin{table*}[t]
    \centering \scriptsize
    \begin{tabular}{lcc}
    \toprule
    Posecode type & Categorization & Condition \\
    \midrule
    \multirow{6}{*}{angle}
        & \texttt{completely bent} & $v \pm 5 \leq 45$ \\
         & \texttt{almost completely bent} &     $45 < v \pm 5 \leq 75$  \\
       &  \texttt{bent at right angle} &   $75 < v \pm 5 \leq 105$  \\
        & \texttt{partially bent} &     $105 < v \pm 5 \leq 135$ \\
        & \texttt{slightly bent} & $135 < v \pm 5 \leq 160$ \\
        & \texttt{straight} & $v \pm 5 > 160$ \\
    \midrule 
    \multirow{4}{*}{distance}
       & \texttt{close} &  $v \pm 0.05 \leq 0.20$ \\
        &\texttt{shoulder width apart} & $0.20 < v \pm 0.05 \le 0.40$ \\
       & \texttt{spread} &  $0.40 < v \pm 0.05 \le 0.80$ \\
       & \texttt{wide} & $v \pm 0.05 > 0.80$ \\
       \midrule
    \multirow{3}{*}{relative position along the X axis}
        & \texttt{at the right of} & $v \pm 0.05 \leq -0.15 $\\
        & \texttt{x-ignored} & $-0.15 < v \pm 0.05 \leq 0.15 $\\
        & \texttt{at the left of} & $v \pm 0.05 > -0.15 $\\
    \midrule 
    \multirow{3}{*}{relative position along the Y axis}
        &\texttt{below}            & $v \pm 0.05 \leq -0.15 $\\
        &\texttt{y-ignored}  & $-0.15 < v \pm 0.05 \leq 0.15 $\\
        &\texttt{above}            & $v \pm 0.05 > -0.15 $\\
    \midrule
    \multirow{3}{*}{relative position along the Z axis}
       & \texttt{behind}                  & $v \pm 0.05 \leq -0.15 $\\
        &\texttt{z-ignored}         & $-0.15 < v \pm 0.05 \leq 0.15 $\\
        &\texttt{in front of}                   & $v \pm 0.05 > -0.15 $\\
    \midrule
    \multirow{3}{*}{pitch \& roll}
        & \texttt{vertical}               & $v \pm 5 \leq 10 $\\
        & \texttt{ignored}       & $10 < v \pm 5 \leq 80 $\\
        & \texttt{horizontal}             & $v \pm 5 > 80 $\\
    \midrule
    \multirow{2}{*}{ground-contact}
        & \texttt{on the ground} & $v \pm 0.05 \leq 0.10$ \\
        & \texttt{ground-ignored} & $v \pm 0.05 > 0.10$ \\
    \midrule
    \multirow{7}{*}{orientation along the X axis}
        & \texttt{handstand} & $v \pm 5 \leq -120$ \\
        & \texttt{lie backward} & $ -120 < v \pm 5 \leq -80$ \\
        & \texttt{lean backward} & $ -80 < v \pm 5 \leq -30$ \\
        & \texttt{orix-organized} & $ -30 < v \pm 5 \leq 30$ \\
        & \texttt{lean forward} & $ 30 < v \pm 5 \leq 80$ \\
        & \texttt{lie forward} & $ 80 < v \pm 5 \leq 120$ \\
        & \texttt{backflip} & $ v \pm 5 > 120$ \\
    \midrule
    \multirow{7}{*}{orientation along the Y axis}
        & \texttt{turn back from right} & $v \pm 5 \leq -150$ \\
        & \texttt{turn clockwise} & $ -150 < v \pm 5 \leq -80$ \\
        & \texttt{slightly turn clockwise} & $ -80 < v \pm 5 \leq -30$ \\
        & \texttt{oriy-organized} & $ -30 < v \pm 5 \leq 30$ \\
        & \texttt{slightly turn counter-clockwise} & $ 30 < v \pm 5 \leq 80$ \\
        & \texttt{turn counter-clockwise} & $ 80 < v \pm 5 \leq 150$ \\
        & \texttt{turn back from right} & $ v \pm 5 > 150$ \\
    \midrule
    \multirow{7}{*}{orientation along the Z axis}
        & \texttt{lie on the right} & $v \pm 5 \leq -80$ \\
        & \texttt{lean right} & $ -80 < v \pm 5 \leq -45$ \\
        & \texttt{slightly lean right} & $ -45 < v \pm 5 \leq -20$ \\
        & \texttt{oriz-ignored} & $ -20 < v \pm 5 \leq 20$ \\
        & \texttt{slightly lean left} & $ 20 < v \pm 5 \leq 45$ \\
        & \texttt{lean left} & $ 45 < v \pm 5 \leq 80$ \\
        & \texttt{lie on the left} & $ v \pm 5 > 80$ \\
    \midrule
    \multirow{3}{*}{translation along the X axis}
        & \texttt{move right} & $v \pm 0.05 \leq -0.3$ \\
        & \texttt{transx-ignored} & $ -0.3 < v \pm 0.05 \leq 0.3$ \\
        & \texttt{move left} & $ v \pm 0.05 > 0.3$ \\
    \midrule
    \multirow{3}{*}{translation along the Y axis}
        & \texttt{squat down} & $v \pm 0.05 \leq -0.2$ \\
        & \texttt{transy-ignored} & $ -0.2 < v \pm 0.05 \leq 0.2$ \\
        & \texttt{jump up} & $ v \pm 0.05 > 0.2$ \\
    \midrule
    \multirow{3}{*}{translation along the Z axis}
        & \texttt{go backward} & $v \pm 0.05 \leq -0.5$ \\
        & \texttt{transz-ignored} & $ -0.5 < v \pm 0.05 \leq 0.5$ \\
        & \texttt{go forward} & $ v \pm 0.05 > 0.5$ \\
    \bottomrule
    \vspace{0.1cm}
    \end{tabular} 
    \caption{\textbf{Conditions for posecode categorizations.} The middle column provides the categorization names of each \textit{posecode} and the last column provides the condition, where the angle is measured in degrees and the distance is measured in meters.}
    \label{tab:elementary_posecodes_thresholds}
\end{table*}
    
\subsection{Timecode Definitions}
We provide the details of the time definitions in~\cref{tab:time}, which can be classified as ``start timing'' and ``durations'' for better temporal awareness of the descriptions. For ``start timing'' in~\cref{tab:time}, $v$ represents: 
\begin{equation}
    v = \frac{T_s}{T},
\end{equation}
where $T_s$ is the frame ID of the start time, and $T$ is the frame number of the whole motion sequence. For ``durations'', $v$ represents:
\begin{equation}
    v = \frac{T_e - T_s}{T},
\end{equation}
where $T_e$ and $T_s$ are the frame ID of the end time and start time.

\begin{table}[t]
    \centering \scriptsize
    \begin{tabular}{lcc}
    \toprule
    Time type & Categorization & Condition \\
    \midrule
    \multirow{5}{*}{start timing}
        & \texttt{begin stage} & $v \leq 0.2$ \\
        & \texttt{early stage} & $ 0.2 < v \leq 0.4$ \\
        & \texttt{mid stage} & $ 0.4 < v \leq 0.6$ \\
        & \texttt{late stage} & $ 0.6 < v \leq 0.8$ \\
        & \texttt{final stage} & $ v > 0.8$ \\
    \midrule
    \multirow{4}{*}{duration}
        & \texttt{for a short time} & $v \leq 0.1$ \\
        & \texttt{for a while} & $ 0.1 < v \leq 0.4$ \\
        & \texttt{for a long time} & $ 0.4 < v \leq 0.8$ \\
        & \texttt{for the whole period} & $ v > 0.8$ \\
    \bottomrule
    \vspace{0.1cm}
    \end{tabular} 
    \caption{\textbf{Conditions for timecode categorizations.} The middle column provides the categorization names of each \textit{posecode} and the last column provides the condition, where the angle is measured in degrees and the distance is measured in meters.}
    \label{tab:time}
\end{table}

\subsection{Description Variabilities}
We provide several samples of the description variabilities in~\cref{tab:variabilities}. For each status of the \textit{motioncode}, \textit{start timing}, or \textit{duration}, we randomly select the expressions to introduce variability into the descriptions.
We also add random skip strategies for \textit{motioncodes}, \textit{start timings}, or \textit{durations} to avoid redundancy and increase the diversity.

\begin{table*}[t]
    \centering \scriptsize
    \begin{tabular}{lc}
    \toprule
    Categorization & Variabilities \\
    \midrule
    \multirow{2}{*}{for a short time} 
    & shortly, for a brief period, for a short duration, for a fleeting moment, for a short spell, for a little while\\
    & for a brief interval, for a short stint\\
    \midrule
    \multirow{2}{*}{begin stage} 
    & in the beginning, initially, at the start, at first, in the initial stages, from the beginning, \\
    & in the initial phase, in the initial stage \\
    \midrule
    \multirow{2}{*}{angle swinging} 
    & swinging, swinging continuously, continuously bending and extending, constantly bending and extending \\
    & regularly bending and extending, continually bending and extending\\
    \midrule
    \multirow{2}{*}{leaning forward} 
    & falling forward, pitching forward, toppling forward, tilting forward, lurching forward, \\
    & tipping forward, bowing forward\\
    \bottomrule
    \vspace{0.1cm}
    \end{tabular} 
    \caption{\textbf{Description Variabilities.} For each status of the \textit{motioncode}, \textit{start timing} or \textit{duration}, we define several synonyms and randomly select the expressions.}
    \label{tab:variabilities}
\end{table*}

\section{Licenses}
Our \dataname dataset is distributed under the CC BY-NC-SA (Attribution-NonCommercial-ShareAlike) license.
For each sub-dataset, the users are required to read the original license first, and we would provide the processed data, \ieno, the extracted SMPL~\cite{smpl} results with the approvals from the original institution.
We will provide a brief license description below:

\begin{itemize}
    \item YouTube Action~\cite{liu2009recognizing} is built for non-commercial scientific research purposes. For details of the dataset, please refer to \href{https://www.crcv.ucf.edu/data/UCF_YouTube_Action.php}{https://www.crcv.ucf.edu/data/UCF\_YouTube\_Action.php}.
    \item Olympic Sports~\cite{niebles2010modeling} is built for non-commercial scientific research purposes. For details of the dataset, please refer to \href{http://vision.stanford.edu/Datasets/OlympicSports/}{http://vision.stanford.edu/Datasets/OlympicSports/}.
    \item ASLAN~\cite{kliper2011action} is built for non-commercial scientific research purposes. For details of the dataset, please refer to \href{https://talhassner.github.io/home/projects/ASLAN/ASLAN-main.html}{https://talhassner.github.io/home/projects/ASLAN/ASLAN-main.html}.
    \item HMDB51~\cite{kuehne2011hmdb} is under the license of \href{https://creativecommons.org/licenses/by/4.0/}{CC-BY 4.0}.
    \item UCF101~\cite{soomro2012ucf101} is built for non-commercial scientific research purposes. For details of the dataset, please refer to \href{https://www.crcv.ucf.edu/data/UCF101.php}{https://www.crcv.ucf.edu/data/UCF101.php}.
    \item Penn Action~\cite{zhang2013actemes} is built for non-commercial scientific research purposes. For details of the dataset, please refer to \href{https://github.com/dreamdragon/PennAction}{https://github.com/dreamdragon/PennAction}.
    \item Charades~\cite{charades} is under the license of \href{http://vuchallenge.org/license-charades.txt}{http://vuchallenge.org/license-charades.txt}.
    \item Kinects-400~\cite{kinetics400} is under the license of \href{https://creativecommons.org/licenses/by/4.0/}{CC-BY 4.0}.
    \item AVA~\cite{gu2018ava} is under the license of \href{https://creativecommons.org/licenses/by/4.0/}{CC-BY 4.0}.
    \item NTU RGB+D 120~\cite{nturgbd120} is created for academic research purposes only. For the details of the license, please refer to \href{https://rose1.ntu.edu.sg/dataset/actionRecognition/}{https://rose1.ntu.edu.sg/dataset/actionRecognition/}.
    \item HAA500~\cite{chung2021haa500} is under the license of \href{https://www.cse.ust.hk/haa/LICENSE}{https://www.cse.ust.hk/haa/LICENSE}.
    \item UAV-Human~\cite{li2021uav} is built for non-commercial scientific research purposes. For details of the dataset, please refer to \href{https://sutdcv.github.io/uav-human-web/}{https://sutdcv.github.io/uav-human-web/}.
    \item MultiSports~\cite{li2021multisports} is under the license of \href{https://creativecommons.org/licenses/by-nc/4.0/}{CC-BY-NC 4.0}.
    \item The codes for preprocessing the data, and training models are under the \href{https://opensource.org/license/mit}{MIT License}.
\end{itemize}

{
    \small
    \bibliographystyle{ieeenat_fullname}
    \bibliography{main}
}


\end{document}